\pdfoutput=1

\documentclass[11pt]{article}

\usepackage[final]{ACL2023}

\usepackage{times}
\usepackage{latexsym}

\usepackage[T1]{fontenc}

\usepackage[utf8]{inputenc}

\usepackage{microtype}

\usepackage{inconsolata}

\usepackage{multirow}
\usepackage{hyperref}
\usepackage{times}
\usepackage{latexsym}
\usepackage[utf8]{inputenc}
\usepackage{xcolor,colortbl}
\usepackage{microtype}
\usepackage{url}
\usepackage{longtable}
\usepackage{tikz}
\usepackage{tabularx}
\usepackage{lscape}
\usepackage{hhline}
\usepackage{bbm}
\usepackage{array}
\usepackage{graphicx}
\usepackage{booktabs}
\usepackage{makecell}
\usepackage{amsfonts}       
\usepackage{nicefrac}       
\usepackage{multirow}
\usepackage{caption}
\usepackage{amsmath}
\usepackage{float} 
\usepackage{ragged2e}
\usepackage{breqn}
\usepackage{soul}
\usepackage{tabularx}
\usepackage{graphicx,subcaption} 
\usepackage[T1]{fontenc}

\title{How Good is Zero-Shot MT Evaluation for\\ Low Resource Indian Languages?}

\author{
Anushka Singh$^{1,2}$ \quad
  Ananya B. Sai$^{1,2}$ \quad
  Raj Dabre$^{1,2,3,6}$\\
  \textbf{Ratish Puduppully}$^{4}$ \quad
  \textbf{Anoop Kunchukuttan}$^{1,2,5}$ \quad
  \textbf{Mitesh M. Khapra}$^{1,2}$
\\
\\
 \textsuperscript{1}Nilekani Centre at AI4Bharat \quad \textsuperscript{2}Indian Institute of Technology Madras, India \\
 \textsuperscript{3}National Institute of Information and Communications Technology, Kyoto, Japan\\
 \textsuperscript{4} Institute for Infocomm Research (I2R), A*STAR, Singapore\\
 \textsuperscript{5}Microsoft, India \quad
 \textsuperscript{6}Indian Institute of Technology Bombay, India
}

\begin{document}
\maketitle
\begin{abstract}
While machine translation evaluation has been studied primarily for high-resource languages, there has been a recent interest in evaluation for low-resource languages due to the increasing availability of data and models. 
In this paper, we focus on a zero-shot evaluation setting focusing on low-resource Indian languages, namely Assamese, Kannada, Maithili, and Punjabi. We collect sufficient Multi-Dimensional Quality Metrics (MQM) and Direct Assessment (DA) annotations to create test sets and meta-evaluate a plethora of automatic evaluation metrics. We observe that even for learned metrics, which are known to exhibit zero-shot performance, the Kendall Tau and Pearson correlations with human annotations are only as high as \textbf{0.32} and \textbf{0.45}. Synthetic data approaches show mixed results and overall do not help close the gap by much for these languages. This indicates that there is still a long way to go for low-resource evaluation. The dataset and evaluation metrics are publicly accessible online.\footnote{\url{https://github.com/AI4Bharat/IndicMT-Eval}} 
\end{abstract}


\section{Introduction}

While there has been a meteoric rise in the amount of data and improvements in architectures for machine translation (MT) models \cite{ai4bharat2023indictrans2,costa2022no}, in order to scientifically establish whether the translation quality has improved, it is important to have reliable evaluation metrics. However, most of the evaluation metrics were developed with English and a few select other languages in mind. It has been shown that such metrics do not necessarily generalize to other languages and have to be separately meta-evaluated \citep{sai-b-etal-2023-indicmt,rivera-trigueros-olvera-lobo-2021-building}. The reasons behind this include linguistic aspects that vary across languages, along with factors like the diversity of outputs produced by the models for each language. Such qualitative differences will be exacerbated in low-resource languages due to the prominent reliance on extensive data resources by today's models.

In this work, we delve deeper into the evaluation of low-resource Indian languages, namely Assamese, Maithili, Punjabi, and Kannada, belonging to 2 different language families. Our goal is to establish the reliability of MT evaluation metrics for low-resource languages. To facilitate this, we collect human scores on the candidate translations using the MQM approach \cite{mqm_lommel2014multidimensional}. We make use of 5 large multilingual models and APIs that can output text in these languages to generate candidate translations for evaluation. We then collect 250 annotations per language, amounting to a total of 1000 MQM annotations for low-resource languages.

Using the data we created, we evaluate multiple existing evaluation metrics of different types, both automatic and learned. In the case of learned metrics, since we do not have training data, we leverage data for related Indic languages from \citet{sai-b-etal-2023-indicmt} for fine-tuning and performing zero-shot meta-evaluations. We observe that for these learned metrics, despite studies finding decent to good performance in other languages, there is a huge margin for improvement in evaluating low-resource languages.
We also explore the influence of the base model and synthetic data generation for low-resource languages. 

In summary, our contributions are as follows:
(i) MQM dataset for 4 low resource languages for evaluation
(ii) Meta-evaluation of existing metrics on low-resource languages
(iii) Analysis of potential techniques to improve the metrics, including (a) exposure of metrics to related languages, (b) different base models, and (c) usage of synthetic data. 
We show that evaluation for low-resource languages is still far behind other languages.

\section{Related Work}
The effectiveness of evaluation metrics has been studied for various languages. Most of the existing MT evaluation metrics are typically analyzed for language pairs where English serves as either the source or target language. That has led to several criticism works \cite{Ananthakrishnan2006SomeII_more_blues,callison-burch-etal-2006-evaluating_bleu_in_mt,sacrebleu} followed by improvements. However, models are getting increasingly multilingual and slowly evaluation metrics are being studied for other languages
\citep{sai-b-etal-2023-indicmt,Freitag2021ExpertsEA,rivera-trigueros-olvera-lobo-2021-building,Cahyawijaya2021IndoNLGBA}. Metrics like chrF \citep{popovic-2015-chrf} and chrF++ \citep{DBLP:conf/wmt/Popovic17_chrFpp} were proposed for character-based, morphologically-rich languages. While some of these criteria hold for the languages we consider, there is no publicly available open study of such metrics for the specific case of low resource languages. 
On the other hand, different evaluation metrics are being used to evaluate models in these languages. WMT23 \cite{pal-etal-2023-findings} had a special task track for low resource Indic languages for which BLEU, ChrF, RIBES, TER, and COMET metrics were used apart from human evaluation.
However, to the best of our knowledge, there are no studies analyzing whether these metrics correlate with human judgments or not for these languages. Additionally, there is no publicly available data with human scores to study this. \citet{Mohtashami2023LearningTQ} used synthetic data augmentation to build a BLEURT-like metric for low resource languages. The only Indian language in their set is Punjabi (2k size, not publicly released), which initially had a poor Pearson correlation of 0.184. This was slightly improved to a value of 0.194 when adding synthetic data to their baseline data, although it is still a poor correlation value.


\begin{table*}[t!]
\footnotesize
\small 
\centering
\resizebox{\textwidth}{!}{
\begin{tabular}{l|cc|cc|cc|cc|cc}
\toprule
            & \multicolumn{2}{c|}{\bf Assamese} & \multicolumn{2}{c|}{\bf Maithili} & \multicolumn{2}{c|}{\bf Kannada} & \multicolumn{2}{c|}{\bf Punjabi} & \multicolumn{2}{c}{\bf Average} \\
\multirow{-2}{*}{\bf Metric}      & $\tau$          & $\rho$          & $\tau$          & $\rho$           & $\tau$          & 	$\rho$           & $\tau$          & $\rho$           & $\tau$          & $\rho$        \\ 
\midrule
BLEU 1      & 0.063      & 0.072      & -0.131      & -0.047      & -0.017      & -0.046      & -0.002      & -0.162      & -0.022      & -0.046           \\
BLEU 2      & 0.058      & 0.081      & 0.078      & -0.028      & 0.016      & 0.035      & -0.016      & 0.065      & 0.034      & 0.038          \\
BLEU 3      & 0.020      & 0.036      & -0.028      & -0.072      & 0.111      & 0.061      & -0.055      & 0.023      & 0.012      & 0.012          \\
BLEU 4      & 0.001      & 0.026      & -0.032      & -0.036      & -0.088      & -0.110      & -0.023      & 0.065      & -0.036      & -0.014          \\
SacreBLEU   & 0.075      & 0.104      & 0.199      & 0.265      & 0.103      & 0.155      & 0.098      & 0.154      & 0.119      & 0.170            \\
ROUGE-L     & 0.088      & 0.128      & 0.052      & 0.055      & 0.005      & 0.003      & -0.074      & 0.065      & 0.018      & 0.063         \\
chrF++      & \bf 0.160      & \bf 0.254      &  0.252      &  0.366      & \bf 0.145      & \bf 0.228      &  0.164     & \bf 0.255      & \bf 0.180      & \bf 0.276         \\
TER         & 0.123      & 0.158      & \bf 0.257      & \bf 0.403      & 0.131      & 0.199      & \bf 0.170      & 0.240      & 0.170      & 0.250        \\
\midrule
LASER embs  & 0.097      & 0.191      & 0.119      & \bf 0.306      & 0.139      & 0.275      & 0.036      & 0.042      & 0.098      & 0.204      \\
LabSE embs  & \bf 0.128      & \bf 0.194      & \bf 0.125      & 0.169      & \bf 0.219      & \bf 0.366      &\bf 0.19      &\bf 0.303      &\bf 0.166      & \bf 0.258         \\
\midrule
mBERT       & 0.131      & 0.247      & 0.212      & 0.388      &  0.165      &  0.248      & 0.234      & 0.281      & 0.186      & 0.291           \\
distilmBERT & 0.139      & 0.267      & 0.250      & 0.416      & 0.169      & 0.263      & \bf 0.245      &  0.306      &  0.201      & 0.313     \\
IndicBERT   & 0.199      & 0.290      & 0.235      & 0.389      & \bf0.191      & \bf 0.276      & 0.237      &\bf 0.311      & 0.216      & 0.317         \\
MuRIL       & \bf 0.206      & \bf 0.324      & \bf 0.309      & \bf 0.476      & 0.162      & 0.239      & 0.204      & 0.269      & \bf0.220      & \bf 0.327     \\
\midrule

BLEURT-20   & 0.119      & 0.185      &\bf0.320      &\bf 0.440      & 0.279      & 0.488      & 0.280      & 0.352      & 0.250      & 0.366            \\
COMET-DA      & 0.228      & 0.298      & 0.172      & 0.264      & 0.281      & 0. 390     & 0.300      & 0.358      & 0.245      & 0.328           \\
COMET-MQM      & 0.260      & 0.381     & 0.199      & 0.291     & 0.290      & 0.410      & 0.266      & 0.334      & 0.254      & 0.354          \\
COMET-QE-DA      & 0.290      & 0.340      & 0.080      & 0.070      & 0.300      & 0.450      & 0.270      & 0.330      & 0.235      & 0.298          \\
COMET-QE-MQM      & 0.230      & 0.350      & 0.130      & 0.200      & 0.300      & 0.440      & 0.220      & 0.290      & 0.220      & 0.320          \\
COMET-Kiwi      &\bf 0.344      & 0.475      & 0.115      & 0.129      &\bf 0.371      &\bf 0.514      &\bf 0.322      & \bf0.392      & 0.288      & 0.378          \\
COMET-Kiwi-xl      & 0.334      &\bf 0.48      & 0.300      & 0.338      & 0.337      & 0.486      & 0.266      & 0.352      & \bf0.309      & \bf0.414          \\

\midrule
GEMBA-MQM      & \bf0.235      & 0.266      &\bf 0.085      &\bf 0.118      & \bf0.108      & \bf0.079     & \bf0.282      & 0.235      & \bf0.178      & \bf0.174           \\
GEMBA-MQM(IL lang)      & 0.228      &\bf 0.276     & 0.081      & 0.077     & 0.050      & 0.069      & 0.171      &\bf 0.261      & 0.132     & 0.171        \\



\midrule
\midrule
 Indic-COMET-DA   & 0.263      & 0.348      & 0.221      & 0.300      & 0.353      & 0.511      & \bf0.293      & \bf0.361      & 0.283      & 0.380           \\
 Indic-COMET-MQM    & 0.201     & 0.270     & 0.201      & 0.288      & 0.251      & 0.388      & 0.282      & 0.340      & 0.234      & 0.322         \\
Base-IndicBERT(DA)      &  0.273      &  0.396      & \bf 0.380      &  \bf0.552      & \bf 0.384      & \bf 0.528      &  0.259     &  0.353      & \bf 0.324      & \bf 0.457         \\
Base-IndicBERT(MQM)         & \bf0.293      & \bf0.426      & 0.311      & 0.483      & 0.302      & 0.440      & 0.224      & 0.313      & 0.283      & 0.416        \\

\midrule
Single Stage  & 0.232  & \bf0.348    & \bf0.337   & \bf0.473   & 0.279    & 0.437   & \bf0.305    & \bf0.378     & \bf0.288      & \bf0.409           \\
2-Stage S/R &  \bf0.234  &  0.345    & 0.264   & 0.360    & \bf0.325    & \bf0.497   & 0.297    & 0.377     & 0.280      & 0.395           \\
2-Stage R/S   & 0.194  & 0.292   & 0.211   & 0.322    & 0.325    & 0.463   & 0.279    & 0.342     & 0.252   & 0.355          \\

\bottomrule
\end{tabular}
}
\caption{Kendall tau ($\tau$)  and Pearson ($\rho$) correlations of various evaluation metrics with human judgements at the segment-level.
The best metric correlation among each category of metrics in \textbf{bold} in the respective block.
The blocks delineate the following categories  (i) word or character overlap-based metrics, (ii) embedding-based metrics, (iii) BERTscore-based formulations with embeddings from different multilingual models, (iv) trained metrics, and (v) GPT-4 based evaluation methods. The blocks after this show the results of our experiments with (a) Finetuning on related languages. (These experiments were done by varying seed values across 5 different runs and the standard deviation to be of the order of 10^-3) (b) adding synthetic data to the training. }
\label{tab:corr_formula_metrics}
\end{table*}
\section{Methodology}
We collect MQM annotations as well as direct assessment (DA) scores and also create synthetic data for 4 languages, viz., Assamese, Punjabi, Kannada, and Maithili.
We use the human-curated data as test data to benchmark the performance of various metrics on these low resource languages. The synthetic data is used to investigate the use of such strategies for augmenting resources in these languages for potential improvements in performance. We design experiments to understand the role of other related languages and the base model on the performance. The following subsections provide the details of the data we create and the strategies explored in our experiments.
\subsection{MQM Data Annotation}
Following \citet{sai-b-etal-2023-indicmt}, for each of the 4 languages, we hired 2 language experts who are native speakers of that language with bilingual proficiency in English. We provided them the English source segment, the translation to be evaluated, and the MQM annotation guidelines \citep{mqm_lommel2014multidimensional, sai-b-etal-2023-indicmt} for identifying error types and their severities in the translations. These annotations were later used to calculate MQM scores. In addition to identifying errors, the annotators were also asked to assign a score to the translation in the range of 0-25, which we refer to as DA score since these are directly assigned by the annotator.\par
For quality assurance, we initially gave 50 common segments to both annotators to mark the errors and indicate their scores. For any disagreements in annotations, the reasons were independently discussed with the annotators. Most of these disagreements were slight differences in marking severity, which we found to be subjective and difficult to standardize. Later, we computed the inter-annotator agreements (IAA) using the Pearson correlation of their scores. We employed a different annotator and repeated the validation process whenever this was below a threshold of 0.5 (which was the case for one language in our set - Punjabi). The final IAA is as follows for the 4 languages considered - Maithili - 0.7, Punjabi - 0.7, Assamese - 0.65, and Kannada - 0.68.  \par
We obtained the translations from 5 state-of-the-art multilingual models and APIs including 
IndicTrans \citep{10.1162/tacl_a_00452}, NLLB  \citep{costa2022no}, NLLB-MoE\footnote{We use the 1.3 B parameter versions of the NLLB models.}, Microsoft Azure Cognitive Services API 
\footnote{\href{https://learn.microsoft.com/en-us/azure/cognitive-services/translator/reference/v3-0-reference}{Bing API}} 
and Google translation API\footnote{\href{https://cloud.google.com/translate/docs/reference/rpc/google.cloud.translate.v2}{Google API}}.
The source segments fed to these models are sampled from the FLORES-101 dataset \citep{goyal-etal-2022-flores}, and each segment is translated by each of the 5 models. These sources and translated segments are presented to the language expert in a random order without details regarding the model / API that generated the translation. The language expert is asked to highlight the text containing the error and indicate the type and severity of the error. We obtain such detailed annotations on 250 segments per language.


\subsection{Synthetic Data Creation}
\label{subsec:synthetic_data}
As human annotation data is expensive and time-consuming to collect, we follow \citet{Geng2023UnifyWA} and \citet{Geng2022NJUNLPsPF} and generate synthetic data for the aforementioned languages to reflect the variety of error types and severities in translations. 
Since we only have test sets, we obtain error type and severity distributions from datasets of related Indic languages in \citet{sai-b-etal-2023-indicmt}. We generate similar proportions of the error types and severities that can be both synthetically 
recreated and have a significant occurrence count
in the distribution. To generate synthetic examples, we utilized BPCC-seed dataset containing data in all these languages without any overlap with the FLORES test set. 
More details about the synthetic data creation are presented in the Appendix \ref{sec:appendix_synthetic}.
Specifically, we created synthetic data with around 44k sentences for Assamese, 32k for Kannada, 24k for Maithili, and 6k for Punjabi based on the size of the available data in these languages.

\subsection{Evaluation Metrics Considered}
We investigate the performance of multiple metrics of different categories. We consider (i) Word-overlap based metrics of BLEU \cite{bleu} variants, SacreBLEU \cite{sacrebleu}, ROUGE \cite{rouge}, (ii) Character-based metric of chrF++ \cite{DBLP:conf/wmt/Popovic17_chrFpp}, (iii) Edit-distance based metric of TER \cite{Snover06astudy_ter}, (iv) Embedding-based metrics of LabSE \cite{feng-etal-2022-language}, LASER \cite{laser} (v) BERTScore computed using mBERT \cite{DBLP:conf/iclr/ZhangKWWA20_BERTscore}, IndicBERT \cite{indicbert} and MuRIL \cite{muril}, (vi) Trained metrics of BLEURT \cite{DBLP:conf/acl/SellamDP20_bleurt} and COMET variants \citep{rei-etal-2020-comet}. Additionally, we also assess GEMBA-MQM \citep{kocmi-federmann-2023-gemba}, a GPT-based reference-free evaluation metric. We experiment with replacing the English-focused examples in the prompt with examples from various Indian languages. We do this by selecting samples in en-hi, en-ta, and en-gu directions from the Indic MT Eval dataset. Further details on our adaptation are provided in Appendix \ref{sec:res_gpt4}.

\subsection{Zero Shot-Evaluation Approach}
\label{sec:zsandsynth}
Since the focus of this paper is zero-shot evaluation of our languages of interest, for learned metrics like COMET-DA and COMET-MQM, we leverage training data containing MQM and DA annotations, for all 5 related Indic languages, henceforth called \textit{related} data, from \citet{sai-b-etal-2023-indicmt}. The related languages include Hindi, Gujarati, Marathi belonging to Indo-aryan family and Tamil and Malayalam belonging to Dravidian language family. There are 1,476 annotated examples in total per language which we split into train, validation and test set containing 1000, 200 and 276 examples respectively for each language. We found that some of the references were mismatched with the source sentences, which we corrected for our fine-tuning experiments.  The validation data is used for early stopping, and the models performing best on the 5 related languages are used for zero-shot evaluation on our 4 languages of interest. We consider fine-tuning existing COMET-DA and COMET-MQM models which are language agnostic and compare them against fine-tuned variants using IndicBERTv2 
\cite{Doddapaneni2022TowardsLN} which is Indic focused. 
Note that XLM-Roberta and hence COMET models has 24 layers while IndicBERT v2 has 12 layers making the latter efficient.\\
\noindent\textbf{Using synthetic data:} Regarding the use of synthetic data, created as described in section~\ref{subsec:synthetic_data}, henceforth called \textit{synthetic}, we consider the following configurations on COMET-DA:
\begin{enumerate}
    \item Single Stage: jointly-trained model on a randomly shuffled mix of \textit{related} data and \textit{synthetic} data.
    \item 2-Stage S/R: training on \textit{synthetic} data followed by \textit{related} data with a reduced learning rate.
    \item 2-Stage R/S: training on \textit{related} data followed by \textit{synthetic} data with a reduced learning rate.
\end{enumerate}

\section{Results}


We present the results for the following research questions to find ways to potentially improve performance on these models: \\
(RQ0) How do existing metrics fare on low-resource languages? 
\\
(RQ1) Does fine-tuning on related languages help?
\\
(RQ2) Does replacing the underlying model of a trained evaluation metric with an alternate backbone model trained on related languages help?
\\
(RQ3) Does synthetic training data help? We report Kendall-tau and Pearson correlations with human annotations.

\subsection{Meta-Evaluation of Existing Metrics}
Table \ref{tab:corr_formula_metrics} shows that, among the word or character overlap-based metrics, chrF++ performs the best on most of the languages. In the embedding-based approach, we find LabSE performs better than LASER embeddings. However, overall the word-based, character-based, and embedding-based metrics are outperformed by the trained metrics.
Among the trained metrics, the COMET model variants perform the best. Specifically, the recently proposed referenceless COMET-Kiwi and COMET-Kiwi-xl models have the best correlations with human judgments. 
However, most of the COMET-variants, except for COMET-Kiwi-xl, perform poorly in the Maithili language. This is despite the COMET*-DA variants having seen Hindi language data during training, which is closely related to Maithili and shares the same script.
We observe that the GPT-4 based evaluation exhibited significantly lower performance on these languages. This could be attributed to the limited exposure of the underlying model to Indian languages, potentially hindering its ability to effectively identify translation errors in this context.
All the analysis above presents observations of the relatively better performing metrics. Overall, we find that none of the evaluation metrics have good correlations with human judgments on these low resource languages.

\subsection{Impact of Related languages}
\label{sec:res_rel_lang_ft}


In the 6th block of Table \ref{tab:corr_formula_metrics}, specifically in the first two rows, we observe that fine-tuning on the 5 related languages improves correlations with human judgments(detailed results in Table \ref{tab:indic_mt_eval_metrics} of Appendix \ref{sec:Appendix_training}).
We find that it also enhances performance of COMET-DA ("Indic-COMET-DA" row) on the low-resource languages belonging to the same or a close language family. 
However, we did not observe the same trend for COMET-MQM ("Indic-COMET-MQM" row).

Our findings suggest that fine-tuning on related languages using supervised data can be a promising technique for improving performance on low resource languages. However, its effectiveness may vary depending on the underlying model and training configuration.

\subsubsection{Does the Backbone Model Matter?}
\label{subsubsec_backbone}
To assess the role of the backbone model on the zero-shot performance, we perform experiments by replacing the XLM-Roberta base model of COMET with the IndicBERT v2 model\footnote{Note that XLM-Roberta model has 24 layers while IndicBERT v2 has 12 layers}. The IndicBERT v2 model is a pretrained multilingual masked language model that was trained on 23 Indian languages including the low-resource languages in our evaluation set. However, note that it used a different dataset namely IndicCorp v2 for training.\par
The rows of `Base-IndicBert(DA)' and `Base-IndicBert(MQM)" in 6th block of Table \ref{tab:corr_formula_metrics}, show what happens when we switch from the COMET backbone to IndicBERT v2. Comparing with non-fine-tuned as well as fine-tuned COMET variants, latter being Indic-COMET, we find that fine-tuning with an Indic-languages-specific base model like IndicBERT v2, which has prior exposure to these languages, leads to an improvement in performance.

\subsection{Training with Synthetic Data}
\label{sec:res_syn_data}
Following the synthetic data incorporation methods outlined in \ref{sec:zsandsynth}, we experiment with using different proportions of the synthetic data with the real data. In particular, we start by adding equal proportions of real and synthetic data (i.e., 5k samples each) and thereafter double the amount of synthetic data added until we hit the maximum amount of data available for synthetic data creation. 

The results are presented in 
Table \ref{tab:corr_formula_metrics}(detailed results in Table \ref{tab:synthetic_result_appendix}).Note that the synthetic data portion added in the experiment for each low resource language only contains data in that particular language. However, the real data consists of the same 5 related Indian languages. 

None of these approaches conclusively outperform the baseline models (COMET-DA and Indic-COMET-DA). The Single-Stage approach shows modest improvement when equal proportions of real and synthetic data are used. However, the performance declines on adding more amount of synthetic data. Overall, the mixed results in these experiments question the effectiveness of using larger quantities of synthetic data for low resource language translation evaluation tasks. This highlights the need for further investigation in this area, presenting an avenue for future research.

\section{Conclusions}
Our work introduced an MQM dataset for four low resource languages consisting of 250 examples per language. Using this dataset, we analyzed the zero-shot performance of different types of existing metrics and observed that none of these existing methods showed good results in the case of low resource language. We explored different techniques to improve the performance, which includes finetuning on related language using Indic MT eval dataset \ref{sec:res_rel_lang_ft}, changing the base model to an Indic-model \ref{subsubsec_backbone} and using synthetic dataset of these low resource language \ref{sec:res_syn_data}. While some of these techniques provide small improvements, we find that there is still a long way to go for low-resource language evaluation.

\section{Limitations}
The size of our dataset being small makes it just about sufficient for testing purposes. The lack of a dev split for the data limits the possibilities of exploring certain other recipes for training. We hope this serves as a starting point though.

\section{Ethical Consideration}
For human annotations, language experts were provided with monthly salary based on their skill set and experience, under the norms of the government of our country. The annotations are collected on a publicly available dataset and will be released publicly for future use. All the datasets created as part of this work will be released under a CC-0 license\footnote{\url{https://creativecommons.org/publicdomain/zero/1.0}} and all the code and models will be released under an MIT license\footnote{\url{https://opensource.org/licenses/MIT}}.

\section*{Acknowledgements}
We would like to express our gratitude to the Ministry of Electronics and Information Technology (MeitY), Government of India, for setting up the ambitious Digital India Bhashini Mission with the goal of advancing Indian language
technology. The annotators and language experts
who worked on this project were supported by the
generous grant given by Digital India Bhashini Mission to IIT Madras to serve as the Data Management Unit for the mission.
We thank Shri Nandan Nilekani and Shrimati Rohini Nilekani for
supporting our work through generous grants from
EkStep Foundation and Nilekani Philanthropies.
These grants were used for (i) supporting many
of the students, research associates, and developers
who worked on this project, (ii) fulfilling many of
our computing needs, and (iii) recruiting project managers to oversee the massive pan-India data collection activity undertaken as a part of this work. We thank Pranjal Agadh Chitale for his helpful feedback and research discussions on this work.
\bibliography{anthology,custom}

\appendix
\section{GPT-4 as evaluator}
\label{sec:res_gpt4}
\begin{table*}[t!]
\footnotesize
\small 
\centering
\resizebox{\textwidth}{!}{
\begin{tabular}{l|cc|cc|cc|cc|cc}
\toprule
             & \multicolumn{2}{c|}{\bf asm} & \multicolumn{2}{c|}{\bf mai} & \multicolumn{2}{c|}{\bf kan} & \multicolumn{2}{c|}{\bf pan} & \multicolumn{2}{c|}{\bf Average} \\
\multirow{-2}{*} {\bf Metric}      & $\tau$          & $\rho$          & $\tau$          & $\rho$           & $\tau$          & 	$\rho$           & $\tau$          & $\rho$           & $\tau$          & $\rho$        \\ 
\midrule
 COMET-DA      & 0.228      & 0.298      & 0.172      & 0.264      & 0.281      & 0. 390     & 0.300      & 0.358      & 0.245      & 0.328           \\
Indic-COMET-DA   & 0.263      & 0.348      & 0.221      & 0.3      & 0.353      & 0.511      & 0.293      & 0.361      & 0.283      & 0.38           \\

\midrule
Stage-1 x  & 0.232  & 0.348    & 0.337   & 0.473   & 0.279    & 0.437   & 0.305    & 0.378     & 0.288      & 0.409           \\
Stage-1 2*x &  0.242 & 0.367    & 0.333   & 0.472    & 0.233    & 0.368   &   -    &  -     & -      & -           \\
Stage-1 4*x & 0.196  & 0.293    & 0.336   & 0.425    & 0.232    & 0.358   &   -    &  -     & -      &  -          \\
\midrule

Stage-2 S/R-x &   0.234  &  0.345    & 0.264   & 0.360    & 0.325    & 0.497   & 0.297    & 0.377     & 0.280      & 0.395           \\
Stage-2 S/R-2*x  &   0.248  &   0.355    & 0.278   & 0.384    & 0.32 &  0.504   &   -    &  -     & -      & -           \\
Stage-2 S/R-4*x  &  0.265  & 0.381    & 0.300   & 0.429    & 0.308    & 0.485   &   -    &  -     & -      & -     \\
\midrule

Stage-2 R/S-x   & 0.194  & 0.292   & 0.211   & 0.322    & 0.325    & 0.463   & 0.279    & 0.342     & 0.252   & 0.355          \\
Stage-2 R/S-2*x & 0.160  & 0.251    & 0.225   & 0.345   & 0.316   &  0.442   &   -    &  -     & -      & -      \\
Stage-2 R/S-4*x & 0.167  & 0.252    & 0.206   & 0.303    & 0.335    & 0.410   &   -    &  -     & -      & -           \\

\bottomrule
\end{tabular}
}
\caption{
KendallTau ($\tau$) and Pearson($\rho$) correlation scores of experiments with synthetic data. First block consists of COMET-DA and Indic-COMET-DA models, followed by the results of different stages varying amount of synthetic data added. Here, x =5000 means 5K examples of synthetic data is added in fine-tuning process. For example, Stage-2 S/R-4*x shows the stage-2 result of a particular language, in which COMET-DA is first fine-tuned on synthetic data of size 4*x i.e. 20k, followed by real data.  }
\label{tab:synthetic_result_appendix}
\end{table*}


We used GEMBA-MQM \citep{kocmi-federmann-2023-gemba}, a GPT-based reference-free evaluation metric. This method employs a "three-shot prompting" technique, where GPT-4 is given three predetermined examples in en-de, en-cs, zh-en language pairs to help it understand the task of identifying error-prone segments in the translated text.
We experiment with replacing the English-focused examples in the prompt with examples from various Indian languages. We selected samples in en-hi, en-ta, en-gu directions by sampling from the Indic MT Eval dataset \citep{sai-b-etal-2023-indicmt}. Table \ref{tab:corr_formula_metrics} shows results both on the vanilla GEMBA-MQM and our modified version named GEMBA-MQM(IL) which explicitly includes Indian language examples within the prompt. 

\section{Synthetic Data Creation}
\label{sec:appendix_synthetic}

We first studied the MQM annotations in the related languages of Hindi, Marathi, Gujarati, Tamil, and Malayalam released by \citet{sai-b-etal-2023-indicmt}. We extracted the counts of various error types with their corresponding severity counts. We choose the error types and severities that can both be synthetically recreated and have a significant occurrence count in the distribution. To recreate the errors in the low resource languages considered in our work, we use the BPCC-seed dataset containing data in all these languages without any overlap with the FLORES test set. 
For each of the error types, we modify correct sentences in the following ways. 
\begin{itemize}
    \item \textbf{Omission errors:} We first determine whether the words are stop words or not depending on their frequency of occurrence in the BPCC corpus. We heuristically determine the top 100 words as the common words or stop words. We randomly drop an uncommon word in each segment sampled for an omission error introduction.
    \item \textbf{Addition errors:} We randomly sample an uncommon word to be introduced at a random position in the segment. We found these errors to be less frequent and accordingly sampled fewer segments to include such errors.
    \item \textbf{Mistranslation errors:} We randomly select tokens to be replaced with a ['MASK'] token. We then sample perturbations using Muril model. To replicate errors of different severities, we sample tokens with reduced generation probabilities to represent more severe pseudo errors. For the generation of varied pseudo translations, we employ a random selection process wherein one token is chosen from the top k tokens with the highest generation probability. Specifically, we set k values at 2,3,5,8 and 10 for different levels of severities.
    \item \textbf{Grammatical errors:} We add, drop, or edit the common words to create fluency-based errors in the segments.
\end{itemize}


\section{Training Details}
\label{sec:Appendix_training}
For training, we follow a similar process as \cite{rei-etal-2020-comet}. We start by loading the encoder initialized with either COMET-DA, COMET-MQM, or IndicBERT weights. We divide our model parameters into two groups: the regressor parameter, which involves the parameters of top feed-forward added for regression, and the encoder parameter, which comprises parameters of the pre-trained encoder. In the initial epoch, the encoder is frozen and only feed-forward is trained with a specific learning rate, after that entire model is trained using different learning rate. For detailed information about hyperparameters, please refer to table \ref{tab:hyperparams_appendix}.

All our experiments used a single RTX 3090 Ti GPU, with a cumulative computational time of 8 hours.
Different experiments in this paper used different learning rates (lr) based on hyperparameter tuning. For fine-tuning Indic-COMET-DA and Indic-COMET-MQM, we found 1.0e-06 and 1.5e-06 learning rates to be the best respectively. While we fine-tuned indicBERT with a slightly higher learning rate of 1.0e-05. 

\begin{table}[t!]
\normalsize
\centering
\begin{tabular}{l c c}
 Hyperparameters     & Values     \\ 
\midrule
batch size & 8\\
loss & mse\\
no. of frozen epochs & 1\\
dropout & 0.1\\
encoder learning rate & 1.0e-06\\
encoder model & XLM-RoBERTa\\
hidden sizes & 3072, 1024\\
layer & mix\\
layerwise decay & 0.95\\
learning rate & 1.5e-05\\
optimizer & AdamW\\
pool & avg\\
\bottomrule
\end{tabular}

\caption{Hyperparameters used to fine-tune Indic-Comet-DA. Note that for different experiments the value of \textit{encoder learning rate} and \textit{learning rate} will change.}
\label{tab:hyperparams_appendix}

\end{table}
\begin{table*}[tbh!]
\footnotesize
\small 
\centering
\resizebox{\textwidth}{!}{
\begin{tabular}{l|cc|cc|cc|cc|cc|cc}
\toprule
            & \multicolumn{2}{c|}{\bf Hindi} & \multicolumn{2}{c|}{\bf Malayalam} & \multicolumn{2}{c|}{\bf Marathi} & \multicolumn{2}{c|}{\bf Tamil} & \multicolumn{2}{c|}{\bf Gujarati} & \multicolumn{2}{c}{\bf Average} \\
\multirow{-2}{*}{\bf Metric}      & $\rho$          & $\tau$           & $\rho$          & $\tau$           & $\rho$          & 	$\tau$           & $\rho$          & $\tau$           & $\rho$          & $\tau$          & $\rho$           & $\tau$           \\ 
\midrule

COMET-DA      & 0.357      &0.457       & 0.516      & 0.707     &0.468      &0.648      &0.539       &0.683       & 0.325      & 0.525      &0.441        & 0.608     \\
COMET-MQM      & 0.432     &0.608     &0.394     &0.301      &0.435      &0.523       &0.504       &0.667       & 0.349      &0.483       &0.423   & 0.516    \\
COMET-QE-DA      & 0.44      & 0.59      & 0.46      & 0.6      & 0.34      & 0.52     & 0.48      & 0.64      & 0.42      & 0.57      & 0.428       & 0.584      \\
COMET-QE-MQM      & 0.45      & 0.64      & 0.34      & 0.44     & 0.29      & 0.4      & 0.5      & 0.67      & 0.38      & 0.43     & 0.392       & 0.516      \\

\midrule
Indic-COMET-DA   & 0.389     & 0.555      & 0.561      & 0.745      & 0.494      & 0.672      & 0.568      & 0.747      & 0.344      & 0.530      & 0.471       & 0.65      \\
Indic-COMET-MQM     & 0.485      & 0.681      & 0.472      & 0.349      & 0.519      & 0.635      & 0.522      & 0.676      & 0.412      & 0.569      & 0.482       & 0.582      \\
Base-IndicBERT(DA)    & 0.378      &  0.597      & 0.508      &  0.713      &  0.524      & 0.684      & 0.462      & 0.614      & 0.352      & 0.538      & 0.445       & 0.629      \\
Base-IndicBERT(MQM)        & 0.443      & 0.673      & 0.398      & 0.350      & 0.484      & 0.624      & 0.424      & 0.559      & 0.379      & 0.525      & 0.426       & 0.546      \\
\bottomrule
\end{tabular}
}
\caption{Segment-level Pearson ($\rho$) and Kendall tau ($\tau$) correlations of different metrics on seen languages.}
\label{tab:indic_mt_eval_metrics}
\end{table*}


\end{document}